\documentclass[letterpaper, 10 pt, conference]{ieeeconf}  

\IEEEoverridecommandlockouts                              

\usepackage[T1]{fontenc}
\usepackage{booktabs}
\usepackage{graphicx}
\usepackage{xcolor}
\usepackage{makecell}
\usepackage{amsmath, amssymb}
\usepackage{colortbl}
\usepackage{url}

\title{\LARGE \bf
A Light-Weight Framework for Open-Set Object Detection with Decoupled Feature Alignment in Joint Space}

\author{Yonghao He$^{1*\dag}$, Hu Su$^{2*\ddagger}$, Haiyong Yu$^{1*}$, Cong Yang$^{3}$, Wei Sui$^{1}$, Cong Wang$^{1}$, and Song Liu$^{4\ddagger}$
\thanks{$^{*}$These authors contributed equally.}
\thanks{$^{\dag}$Project lead.}
\thanks{$^{\ddagger}$Corresponding author.}
\thanks{$^{1}$D-Robotics}
\thanks{$^{2}$State Key Laboratory of Multimodal Artificial Intelligence Systems (MAIS), Institute of Automation of Chinese Academy of Sciences}
\thanks{$^{3}$BeeLab, School of Future Science and Engineering, Soochow University}
\thanks{$^{4}$the School of Information Science and Technology, ShanghaiTech University}
}

\begin{document}

\maketitle
\thispagestyle{empty}
\pagestyle{empty}

\begin{abstract}

Open-set object detection (OSOD) is highly desirable for robotic manipulation in unstructured environments. However, existing OSOD methods often fail to meet the requirements of robotic applications due to their high computational burden and complex deployment. To address this issue, this paper proposes a light-weight framework called Decoupled OSOD (DOSOD), which is a practical and highly efficient solution to support real-time OSOD tasks in robotic systems. Specifically, DOSOD builds upon the YOLO-World pipeline by integrating a vision-language model (VLM) with a detector. A Multilayer Perceptron (MLP) adaptor is developed to transform text embeddings extracted by the VLM into a joint space, within which the detector learns the region representations of class-agnostic proposals. Cross-modality features are directly aligned in the joint space, avoiding the complex feature interactions and thereby improving computational efficiency. DOSOD operates like a traditional closed-set detector during the testing phase, effectively bridging the gap between closed-set and open-set detection. Compared to the baseline YOLO-World, the proposed DOSOD significantly enhances real-time performance while maintaining comparable accuracy. The slight DOSOD-S model achieves a Fixed AP of $26.7\%$, compared to $26.2\%$ for YOLO-World-v1-S and $22.7\%$ for YOLO-World-v2-S, using similar backbones on the LVIS \texttt{minival} dataset. Meanwhile, the FPS of DOSOD-S is $57.1\%$ higher than YOLO-World-v1-S and $29.6\%$ higher than YOLO-World-v2-S. Meanwhile, we demonstrate that the DOSOD model facilitates the deployment of edge devices. The codes and models are publicly available at \textcolor{blue}{\url{https://github.com/D-Robotics-AI-Lab/DOSOD}}.

\end{abstract}

\section{Introduction}
\label{introduction}


Object detection aims to predict the locations and category labels of objects of interest, serving as a fundamental technique for understanding images and scenes with numerous applications in robotics \cite{c1, c2} and autonomous vehicles \cite{c3}. 
Significant advances in this field have been driven by the advent of convolutional neural networks (CNNs) \cite{c4}.
Previous research focuses primarily on closed-set condition, aiming to detect objects of only predefined categories. 
When faced with unknown objects, the detectors tend to either ignore them or misclassify them as known categories. 
However, for robotic manipulation in unstructured environments, it is essential to detect both known and unknown objects \cite{c5}. 
Open-set object detection (OSOD) offers a potential solution, which requires models to generalize from a base set to novel categories without specific annotations for new categories. 

Dhamija \textit{et al.} \cite{c7} formalized OSOD task and reported that closed-set detectors experience significant performance degradation in open-set conditions. 
Given the strong open-set image recognition capabilities of vision-language models (VLMs) pre-trained with contrastive methods \cite{c6, c8}, considerable efforts have been made to detect arbitrary classes by leveraging language generalization. 
Recent studies \cite{c9, c10, c11, c12, c13, c14} have explored extending the VLM model, Contrastive Language-Image Pre-training (CLIP) \cite{c6}, to detect objects from novel categories. 
Yao \textit{et al.} \cite{c15, c16} unified detection and image-text datasets through region-text matching, training detectors with large-scale image-text data pairs to achieve promising performance. 

\begin{figure*}
    \centering
    \includegraphics[width=7in]{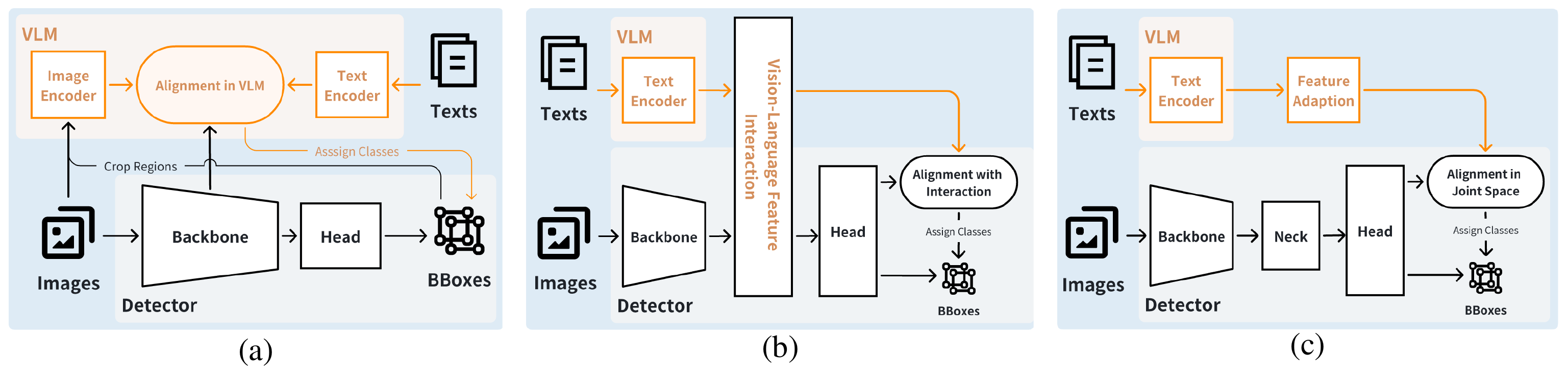}
    \caption{Feature alignment strategy. (a) illustrates the teacher-student distillation approach, where image features extracted by the VLM and the detector are aligned under the supervision of text embeddings generated by the VLM's text encoder. Alternatively, proposals can be used to crop the image, with region features then aligned in a similar manner. (b) shows the interaction-based alignment strategy, where text embeddings interact with image features extracted by the detector's backbone to achieve alignment. (c) presents the proposed decoupled alignment strategy, which aligns features without any interaction.}
    \label{align_comp}
\end{figure*}

However, these methods suffer from \textit{\textbf{a high computational burden and complex deployment}} for the following reasons: (1) \textit{\textbf{heavy backbones}} such as ATSS \cite{c18} or DINO \cite{c19} with Swin-L \cite{c20} employed by the methods in \cite{c14, c15, c16}; (2) \textit{\textbf{utilization of image encoder of VLM}} in \cite{c9, c10, c11, c12, c13} which increases inference cost in testing; (3)\textit{\textbf{ cross-modality feature interaction for alignment}} in \cite{c14, c40}, as illustrated in Fig.~\ref{align_comp}. Additionally. the \textit{\textbf{transformer architecture}} used in these methods adds additional complexity to deployment on edge devices. Therefore, the methods still struggle for real-world applications. 
The recent work, YOLO-World \cite{c17}, introduces a prompt-then-detect strategy that first encodes a user's prompt to build an offline vocabulary and then performs inference on this vocabulary without re-encoding the prompts. 
By using a fully CNN detector, YOLOv8 \cite{c29}, YOLO-World achieves a $20 \times$ speed increase compared to previous state-of-the-art (SOTA) methods. 
However, YOLO-World still suffers from complex feature interactions for alignment, which hinders further efficiency improvements.

Inspired by the recent work, YOLO-World, we adopt a similar pipeline and take a further step in cross-modality feature alignment. Actually, feature alignment is a key concern in current OSOD research.
Fig. \ref{align_comp}(a) depicts the teacher-student distillation approach, where features are extracted from the entire image \cite{c9, c10} or proposals \cite{c11, c12, c13} using the image encoder of VLM, which serves as the teacher. 
Meanwhile, the features extracted by the detector's backbone act as the student. 
The student features are distilled from the teacher features by directly using L1 loss between them or/and under the supervision of text embedding generated by the text encoder of VLM.
Fig. \ref{align_comp}(b) illustrates alignment through feature interaction, involving complex cross-attention \cite{c14} or repeated fusion \cite{c15} operations between multi-modal and multi-level features.
 
This paper proposes a simple yet efficient feature alignment strategy to faciliate real-time OSOD, as shown in Fig. \ref{align_comp}(c). A Multilayer Perceptron (MLP) adaptor is introduced to enhance the learning capacity of text embeddings while transforming them into a latent space.
Simultaneously, a detector learns class-agnostic proposals and then the region representations in the latent space. Cross-modality features can then be aligned directly in the latent space in a decoupled manner, reducing both computational costs and storage requirements.
Inspired by concepts in robotics, we refer to this latent space as a "joint space", indicating that it links multi-modal features in a manner similar to how it connects different robotic arms. A high-efficiency OSOD framework called Decoupled OSOD (DOSOD) is thus established.
During the testing phase, DOSOD operates like a traditional closed-set object detector, efficiently bridging the gap between closed-set and open-set object detection. Extensive experiments on common benchmarks demonstrate that, compared to the state-of-the-art baseline, YOLO-World, the proposed DOSOD significantly improves real-time performance while maintaining comparable accuracy.

The contributions of the paper are summarized as follows:
\begin{itemize}
    \item A light-weight OSOD framework, decoupled OSOD (DOSOD), is proposed, exhibiting high efficiency for low-cost applications. DOSOD bridges the gap between open-set and closed-set detection, thus expanding application scenarios.
    
    \item An MLP-based adaptor is introduced, with which feature alignment is conducted in a decoupled manner, improving computational efficiency.
    
    \item Extensive experiments are conducted to demonstrate that the DOSOD framework significantly improves real-time performance without compromising accuracy. 

\end{itemize}

\section{RELATED WORK}

\subsection{Closed-Set Object Detection}

Methods for closed-set object detection are generally categorized into three groups: region-based, pixel-based, and query-based methods. Region-based methods are exemplified by the two-stage R-CNN series \cite{c22, c23} and one-stage detectors \cite{c24}. These methods rely on a set of predefined anchors, refining and classifying them with non-maximum suppression (NMS) as a post-processing step. Pixel-based methods \cite{c25} predict a bounding box and its corresponding category for each pixel on multiple feature maps. Query-based methods \cite{c26, c27} use a fixed, small set of learned object queries to reason about object relationships and the global image context, directly outputting the final set of predictions. \cite{c26} was the first to employ the transformer architecture in object detection, inspiring subsequent query-based methods \cite{c27} that enhance queries with prior knowledge. Redmon \textit{et al.} proposed the YOLO series \cite{c28, c29} to leverage simple architectures and learning strategies for improved real-time performance. Closed-set object detection methods are limited to handling a fixed set of categories and are not practical in unstructured environments where unknown objects may appear alongside known ones. This paper explores the OSOD task, which aims to detect objects beyond predefined categories, thereby extending the application scenarios of object detection.

\subsection{Open-set Object Detection}

Dhamija \textit{et al.} \cite{c6} formalized the OSOD task which has recently become a new trend in object detection. Current studies focus on training detectors using existing bounding box annotations and leveraging language generalization to detect arbitrary categories. Vision and Language Knowledge Distillation (ViLD) \cite{c9, c10} 
distills the knowledge from a pretrained VLM into a two-stage detector. Du \textit{et al.} \cite{c11} proposed learning continuous prompt representations for OSOD, which is combined with ViLD to achieve superior performance.
Subsequently, feature alignment is enhanced through dense-level alignment \cite{c12} or by improving region representation using a bag of regions \cite{c13}. However, these methods incur high computational costs during the testing phase due to the use of VLM image encoders. \cite{c14, c17} used only the text encoder to replace the vanilla classifier with fixed category text embeddings. \cite{c15, c16} directly learned the fine-grained word-region alignment from massive image-text pairs in an end-to-end manner. The detector was trained with a hybrid supervision from detection, grounding and image-text pair data under a unified data formulation.
While the methods \cite{c14, c15, c16} employ heavy backbones to enhance open-set detection capacity, YOLO-World \cite{c17} introduced a prompt-then-detect pipeline with the YOLOv8 \cite{c30} detector to achieve real-time OSOD. Inspired by YOLO-World, this paper proposes a new OSOD framework, Decoupled OSOD, which further improves real-time performance without compromising accuracy.

\subsection{Feature Alignment}

Feature alignment is a significant concern in current OSOD research. The methods in \cite{c9, c10, c11, c12, c13} adopted the teacher-student distillation approach. ViLD \cite{c9, c10} uses an L1 distillation loss to align region embeddings with image embeddings. \cite{c12} proposed a dense-level alignment rather than object-level alignment, while \cite{c13} introduced the idea of aligning the embedding of a bag of regions instead of individual regions. In the methods, entire image or image patches are inputted into the image encoder of VLM and the backbone of the detector for feature extraction, which are then treated as the teacher and student, respectively. The 'student' features are distilled from the 'teacher' features by using the loss between them or with the supervision of the text embedding. \cite{c14} employs the transformer architecture, and uses cross-attention strategy for cross-modality feature fusion. \cite{c17} incorporated the text embedding into the neck part of the detector for alignment. In \cite{c14, c17}, repeated feature fusion and complex cross-attention operations are conducted, which increases the computational cost. In this paper, we endow the learning capacity of text embedding and transform it into the joint space, in which feature alignment is achieved without interaction. The proposed method further improves the efficiency of OSOD task which facilitates low-cost applications in robotics.

\section{Method}
\label{method}


In this section, a thorough analysis is conducted regarding the relationship between closed-set and open-set detectors. On the basis, the proposed DOSOD framework is detailed.

\subsection{Bridge the gap between closed-set and open-set detectors}
\label{bridge_the_gap_between_closed_set_and_open_set_detectors}
We start with a brief overview of closed-set and open-set detection. Closed-set detection is based on a predefined set of fixed categories, each with corresponding annotations provided during the training phase. In the testing phase, the detector is tasked only with predicting these predefined categories while ignoring any others. In contrast, open-set detection significantly broadens the range of detectable categories by using texts as category labels. During training, a large number of text-region pairs are used as training data. In the testing phase, the model can accept any diverse text input to detect new categories of objects that were never encountered during training.
For closed-set detection, instance annotations are denoted as $\mathbf{A}=\{B_i, c_i\}_{i=1}^{N}$, in which $B_i$ are bounding boxes and $c_i$ are corresponding category labels in form of number indexes.
Instance annotations of open-set detection are formulated as $\mathbf{A}=\{B_i, t_i\}_{i=1}^{N}$, where $t_i$ are texts in form of category names, phrases or captions.
Open-set detection introduces a new perspective, that is, regarding labels as a separate modality.

\begin{figure}
    \centering
    \includegraphics[width=3in]{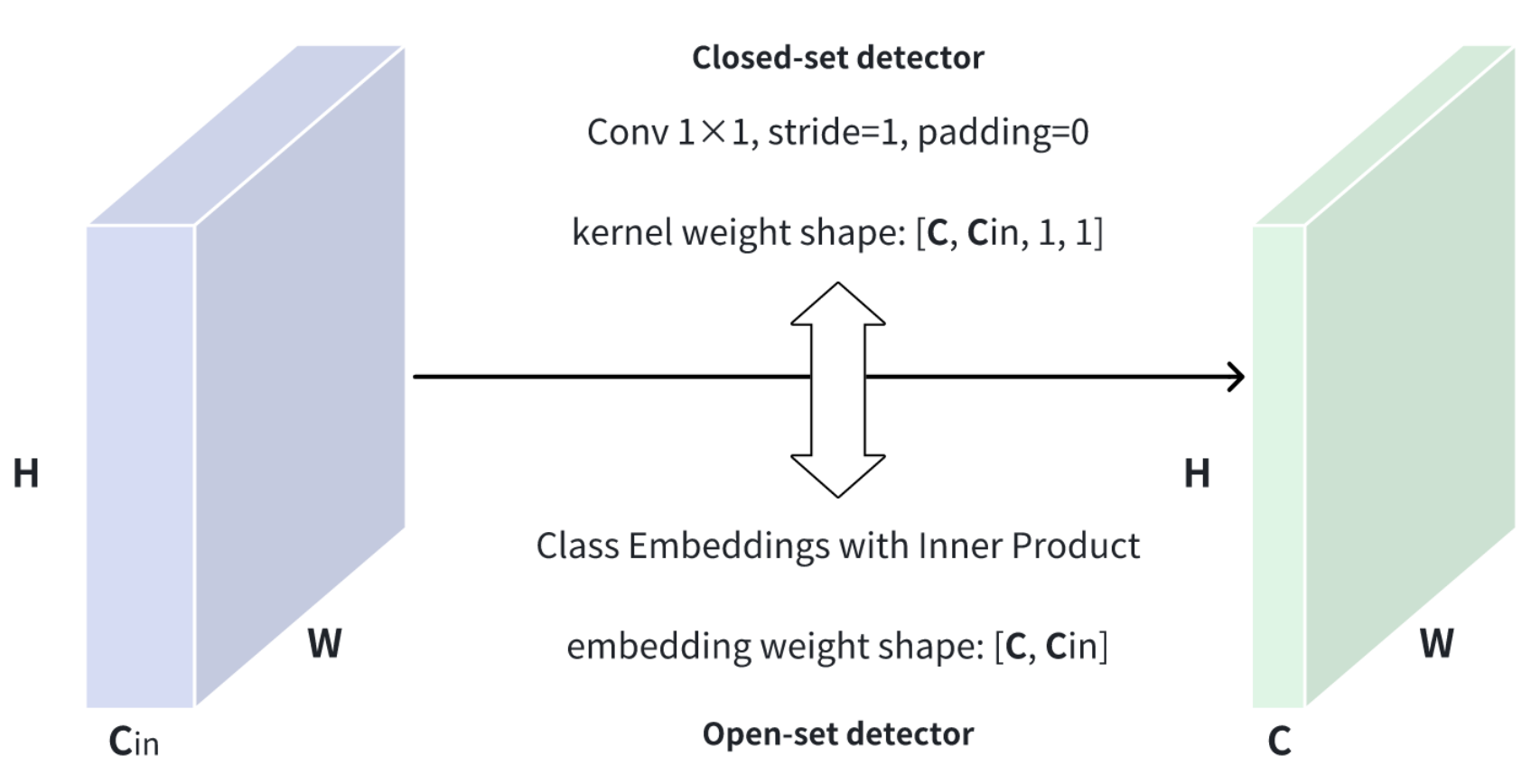}
    \caption{The difference in the last layer of the classification branch between closed-set and open-set detection}
    \label{brige-gap}
    \vspace{-8pt}
\end{figure}


We explain the last layer in the classification branch of the head part to illustrate the difference, as shown in
Fig. \ref{brige-gap}.
In a typical closed-set detector, the last layer of classification branch can be deconstructed equivalently into inner product operation with category embeddings. 
Specifically, each kernel in convolution of the last layer is naturally a randomly-initialized embedding for a certain category (\textit{bias} is omitted for simplicity).
The similarities between category embeddings and image region features are computed for category assignment.
Once the detection problem is regarded as a multi-modality alignment issue, a significant stride is taken from closed-set detection to open-set detection.
Naturally, two principal steps emerge as crucial elements: 1) the number of embeddings can be further increased and their forms are extended to phrases and captions; 
2) learning for alignment between texts and image regions is crucial and should be carefully addressed.
By following the aforementioned steps, we can transform a closed-set detector into an open-set detector, thereby bridging the gap between closed-set and open-set detection paradigms.

\subsection{Model Architecture}
\label{model_architecture}

\begin{figure*}
    \centering
    \includegraphics[width=4.5in]{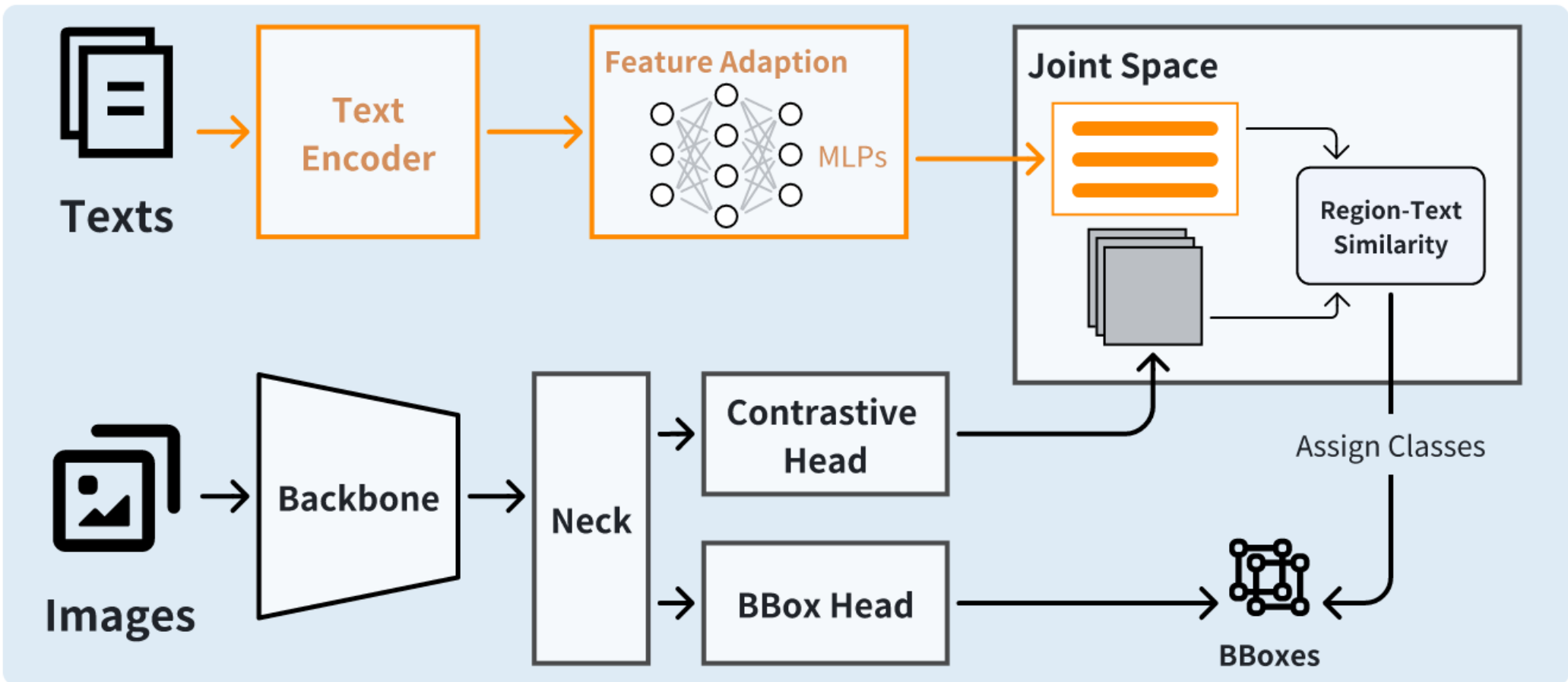}
    \caption{Overview of our DOSOD framework. A detector learns class-agnostic proposals, and the category text embeddings for these proposals are computed using the VLM's text encoder. The embeddings are transformed by the MLP based  adaptor and then aligned with the region features extracted by the detector. The transformed text embeddings serve as the classifier. During inference, text embeddings of novel categories are used to enable zero-shot detection.}
    \label{model_overview}
\end{figure*}

We illustrate the model architecture in Fig.~\ref{model_overview}.
The decoupled architecture explicitly processes texts and images separately, and there is no intermediate interaction between the two modalities.
Specifically, a text encoder is utilized to extract initial embeddings for texts. 
Here, we adopt the text encoder in CLIP \cite{c6} to complete the task.
Subsequently, a text feature adaptor based on MLPs is designed for projecting initial text embeddings to the joint space.
The text feature adaptor is the crucial component for feature alignment in the joint space, instead of feature interaction in the middle of the structure.
This brings simplicity to the structure.
For images, a general single-stage detector with a backbone, neck, and head is employed to extract region features.
The detector is instantiated with YOLOv8 which is used in YOLO-World \cite{c17} as well.
As usual, bbox head predicts class-agnostic bboxes.
Contrastive Head provides region features for alignment in the joint space.
In the joint space, mutual similarities between projected text embeddings and region features are computed to determine the classes that bboxes belong to.
The framework in Fig.~\ref{align_comp}(b) places the feature alignment operation in the middle processing via interactions. 
While the proposed framework places the alignment operation in the final joint space via a text adaptor. 
Hence, the proposed framework can retain the independence of modality-specific structures.

\textbf{MLP Adaptor.} The MLP adaptor transforms text embeddings into the joint space, which can be formulated as:
\begin{equation}
    \mathbf{e}^{\prime} = MLP(\mathbf{e}), 
\end{equation}
where $\mathbf{e}$ is the text embedding, $\mathbf{e}^{\prime}$ is the transformed embedding by the MLP adaptor. The MLP adaptor includes $N$ layers, in which the $i$-th layer can be expressed as:
\begin{equation}
    \mathbf{e}_{i+1} = ReLU(\mathbf{W}_{i}\mathbf{e}_{i} + b_i),
\end{equation}
where $\mathbf{W}_{i}$ and $b_i$ are weight and bias of the linear projection, $ReLU(\cdot)$ is the nonlinear activation of rectified linear unit, $\mathbf{e}_{i}$ and $\mathbf{e}_{i+1}$ are input and output of the $i$-th layer, respectively.
In the proposed architecture, the number of layers is a hyper-parameter. 
All text embeddings share the identical MLPs and the last layer does not undergo the nonlinear activation.

\textbf{Re-parameterization for efficiency.} In the offline vocabulary inference mode, text embeddings can be pre-computed in advance by utilizing the text encoder and text adaptor. 
Subsequently, all these embeddings can be re-parameterized as the kernels of the final convolution:
\begin{align}
    \mathbf{E}^{\prime} = unsqueeze(\mathbf{E}), \\
    \mathbf{X}^{\prime} = Conv_{\mathbf{\theta}=\mathbf{E}^{\prime}}(\mathbf{X}),
\end{align}
where $\mathbf{E} \in \mathbb{R}^{K \times D}$ is pre-computed $K$ text embeddings with $D$-dimensions, $unsqueeze()$ function expands the dimension shape of $\mathbf{E}$ from $K \times D$ to $K \times D \times 1 \times 1$, $Conv()$ operation utilizes $\mathbf{E}^{\prime}$ as kernel weight to process input $\mathbf{X} \in \mathbb{R}^{B \times C \times H \times W}$ and $\mathbf{X}^{\prime} \in \mathbb{R}^{B \times C \times H \times W}$ is the output of $Conv()$.
Thus, the inference efficiency is approximately on a par with that of the individual detector. 
In fact, such an operation transforms the open-set detector into a closed-set detector.




\subsection{Loss Function}
\label{loss_function}
The proposed method fully adopt the loss function from the closed-set detector.
In this paper, we choose YOLOv8 as the detector, and the corresponding loss function is presented as follows:
\begin{equation}
    \mathcal{L} = \lambda_1 \mathcal{L}_{cls} + \lambda_2 \mathcal{L}_{iou} + \lambda_3 \mathcal{L}_{dfl},
\end{equation}
where $\mathcal{L}_{cls}$ is the classification loss based on contrastive similarities, $\mathcal{L}_{iou}$ and $\mathcal{L}_{dfl}$ are losses for bbox regression, $\lambda_1, \lambda_2, \lambda_3$ are loss weights.
The label assignment strategy is the same as that of YOLOv8, namely  task-aligned label assignment \cite{c45}.

\section{Experiments and Results}
\label{experiments_and_results}

\begin{table*}[h]
\caption{Zero-shot Evaluation on LVIS. The methods trained on the same datasets as ours, i.e., o365 and GoldG, are highlighted with a background color for distinction.}
\label{zero_shot_lvis}
\begin{center}
\begin{tabular}{l|llccccc}
\textbf{Method} & \textbf{Backbone} & \textbf{Params} & \textbf{Pre-trained Datasets} & \textbf{AP} & $\mathrm{\textbf{AP}}_{r}$ & $\mathrm{\textbf{AP}}_{c}$ & $\mathrm{\textbf{AP}}_{f}$\\
\hline
MDETR \cite{c34} & ResNet-101 & 169M & GoldG & 24.2 & 20.9 & 24.3 & 24.2 \\ \rowcolor[HTML]{ECF4FF}
GLIP-T \cite{c40} & Swin-T \cite{c20} & 232M & O365, GoldG & 24.9 & 17.7 & 19.5 & 31.0 \\
GLIP-T \cite{c40} & Swin-T \cite{c20} & 232M & O365, GoldG, Cap4M & 26.0 & 20.8 & 21.4 & 31.0 \\ \rowcolor[HTML]{ECF4FF}
GLIPv2-T \cite{c41} & Swin-T \cite{c20} & 232M & O365, GoldG & 26.9 & - & - & - \\ 
GLIPV2-T \cite{c41} & Swin-T \cite{c20} & 232M & O365, GoldG, Cap4M & 29.0 & - & - & - \\ \rowcolor[HTML]{ECF4FF}
Grounding DINO-T \cite{c14} & Swin-T \cite{c20} & 172M & O365, GoldG & 25.6 & 14.4 & 19.6 & 32.2 \\
Grounding DINO-T \cite{c14} & Swin-T \cite{c20} & 172M & O365, GoldG, Cap4M & 27.4 & 18.1 & 23.3 & 32.7 \\
Grounding DINO 1.5 Pro \cite{c42}  & ViT-L \cite{c43} & - & Grounding-20M & 55.7 & 56.1 & 57.5 & 54.1 \\
Grounding DINO 1.5 Edge \cite{c42} & EfficientViT-L1 \cite{c44} & - & Grounding-20M  & 33.5 & 28.0 & 34.3 & 33.9 \\ \rowcolor[HTML]{ECF4FF}
DetCLIP-T \cite{c15} & Swin-T \cite{c20}  & 155M & O365, GoldG & 34.4 & 26.9 & 33.9 & 36.3 \\
\hline \rowcolor[HTML]{ECF4FF}
YOLO-World-v1-S \cite{c17} & YOLOv8-S & 77M & O365, GoldG & 26.2 & 19.1 & 23.6 & 29.8 \\ \rowcolor[HTML]{ECF4FF}
YOLO-World-v1-M \cite{c17} & YOLOv8-M & 92M & O365, GoldG & 31.0 & 23.8 & 29.2 & 33.9 \\ \rowcolor[HTML]{ECF4FF}
YOLO-World-v1-L \cite{c17} & YOLOv8-L & 110M & O365, GoldG & 35.0 & 27.1 & 32.8 & 38.3 \\ \rowcolor[HTML]{ECF4FF}
YOLO-World-v2-S \cite{c17} & YOLOv8-S & 77M & O365, GoldG & 22.7 & 16.3 & 20.8 & 25.5 \\ \rowcolor[HTML]{ECF4FF}
YOLO-World-v2-M \cite{c17} & YOLOv8-M & 92M & O365, GoldG & 30.0 & 25.0 & 27.2 & 33.4 \\ \rowcolor[HTML]{ECF4FF}
YOLO-World-v2-L \cite{c17} & YOLOv8-L & 110M & O365, GoldG & 33.0 & 22.6 & 32.0 & 35.8 \\ \rowcolor[HTML]{ECF4FF}
\hline
DOSOD-S & YOLOv8-S & 75M & O365, GoldG & 26.7 & 19.9 & 25.1 & 29.3 \\ \rowcolor[HTML]{ECF4FF}
DOSOD-M & YOLOv8-M & 90M & O365, GoldG & 31.3 & 25.7 & 29.6 & 33.7 \\ \rowcolor[HTML]{ECF4FF}
DOSOD-L & YOLOv8-L & 108M &O365, GoldG & 34.4 & 29.1 & 32.6 & 36.6 \\
\hline
\end{tabular}
\end{center}
\vspace{-8pt}
\end{table*}

In this section, we conducted extensive experiments of zero-shot detection on benchmark datasets and the results are reported. 
Ablation studies on the text feature adaptor are carried out. 
Finally, the efficiency of our method and that of the counterparts is illustrated.

\subsection{Pre-training Datasets and Implementation Details}
\label{pre-train_datasets_and_implementation_details}
Following \cite{c17}, the datasets employed for pre-training comprise: 1) Objects365v1 \cite{c31} with 609k images and 9,621k annotations; 2) GQA \cite{c32} with 621k images and 3,681k annotations; 3) Flickr30k \cite{c33} with 149k images and 641k annotations. Objects356v1 is created for object detection task, and the category names are treated as texts. GQA and Flickr datasets are merged to form GoldG \cite{c34}. 

The proposed DOSOD method is implemented with code from YOLO-World \cite{c17}, which is based on MMYOLO \cite{c35} and MMDetection \cite{c36}. 
We employ YOLOv8 \cite{c30} as the detector, with multiple versions including small (S), medium (M), and large (L). 
Text embeddings are obtained by utilizing the text encoder of CLIP \cite{c6}. 
The parameters of the text encoder are frozen during training. 
We pre-train DOSOD on 8 NVIDIA RTX 4090 GPUs with a batchsize of 128 while YOLO-World uses 32 NVIDIA V100 GPUs with the batchsize of 512. Note that, we adopt a smaller batchsize in the experiments due to hardware limitations, which might have a negative impact on accuracy to a certain extent. Other settings, such as the optimizer, learning rate, and augmentations, are the same as those of YOLO-World. More details please refer to \cite{c17}.


\subsection{Zero-shot Evaluation}
\label{zero-shot_evaluation}
The datasets used for pre-training include a wide range of categories, phrases, and captions. Once pre-trained, the models are naturally capable of performing zero-shot detection. In this section, we conduct zero-shot evaluations on the LVIS \cite{c34} and COCO \cite{c37} datasets to demonstrate the effectiveness of the proposed method.
We compare our method with the baseline YOLO-World, which has two versions: YOLO-World-v1 and YOLO-World-v2. 
YOLO-World-v2 optimizes the neck and head of the network on the basis of YOLO-World-v1. 
It makes the model more efficient with a slight sacrifice in accuracy.
Therefore, YOLO-World-v2 is the main counterpart baseline.
Additionally, comparisons with other prevalent methods are conducted on the LVIS dataset for further demonstration.



\subsubsection{Results on LVIS Dataset}
\label{results_on_lvis_dataset}
Zero-shot detection is performed on LVIS \texttt{minival}, and the metric of \textit{Fixed AP} \cite{c38} is used for comparison. 
The results are presented in Table \ref{zero_shot_lvis}. 
We found from the table that Grounding DINO 1.5 Pro achieves outstanding performance with a heavy backbone ViT-L \cite{c43} and a huge dataset named Grounding-20M. 
Our method is nearly on a par with YOLO-World-v1 in terms of accuracy but superior to YOLO-World-v2.
Especially, with the YOLOv8-S detector, our DOSOD-S model is marginally better than the YOLO-World-v1-S model ($26.7\%$ $vs$ $26.2\%$) and significantly outperforms the YOLO-World-v2-S model ($26.7\%$ $vs$ $22.7\%$). 
As the model size increases, the accuracy gap between our method and YOLO-World-v2 is gradually narrowing.
It is also noted from the table that, when pre-trained on the same datasets, the light-weight DOSOD-S model outperforms GLIP-T and Grounding DINO-T, and is comparable to GLIPv2-T, all of which use the heavy Swin-T \cite{c20} backbone.
Meanwhile, the DOSOD-L model achieves the highest AP score ($34.4\%$) among models pre-trained on the same datasets with fewer parameters ($108 M$ $vs$ $232 M$).
We believe that, the proposed DOSOD method could be further improved by pre-training on larger-scale datasets.
The results demonstrate that our method has sufficient competitiveness in accuracy.
In Section \ref{efficiency}, we will prove that our method has significant advantages in inference speed.

\subsubsection{Results on COCO Dataset}
\label{results_on_coco_dataset}
We conduct a comparison between the pre-trained DOSOD models and YOLO-World on the COCO dataset for zero-shot detection. 
The objects in COCO is divided into 80 categories, and the category names are treated as texts. 
We directly utilize the models pre-trained on Object365v1 and GoldG for transfer. 
The evaluation metric mAP is computed on COCO \texttt{val2017} for comparison. 
The results are reported in Table \ref{zero_shot_coco}. 
The table shows that DOSOD exhibits slightly inferior performance to YOLO-World-v1 and YOLO-World-v2 by approximately 1\% mAP.
Specially, DOSOD-L is slightly better than YOLO-World-v1-L ($44.6\%$ $vs$ $44.4\%$).
This discrepancy may be attributed to the fact that COCO is a closed-set detection dataset with insufficient text richness, leading to indistinguishable evaluation results.


\begin{table}[h]
\caption{Zero-shot Evaluation on COCO. We report the mAPs of YOLO-World series and our method on COCO val2017. }
\label{zero_shot_coco}
\begin{center}
\begin{tabular}{l|ccccc}

\textbf{Method} & \textbf{AP} & $\mathrm{\textbf{AP}}_{50}$ & $\mathrm{\textbf{AP}}_{75}$ \\
\hline
YOLO-World-v1-S & 37.6 & 52.3 & 40.7 \\
YOLO-World-v1-M & 42.8 & 58.3 & 46.4 \\
YOLO-World-v1-L & 44.4 & 59.8 & 48.3 \\
YOLO-World-v2-S & 37.5 & 52.0 & 40.7 \\
YOLO-World-v2-M & 42.8 & 58.2 & 46.7 \\
YOLO-World-v2-L & 45.4 & 61.0 & 49.4 \\
\hline
DOSOD-S & 36.1 & 51.0 & 39.1 \\
DOSOD-M & 41.7 & 57.1 & 45.2 \\
DOSOD-L & 44.6 & 60.5 & 48.4 \\
\hline
\end{tabular}
\end{center}
\vspace{-8pt}
\end{table}

\subsection{Ablation Study}
\label{ablation_study}
The text feature adaptation is the core component of the proposed method. 
Specifically, we employ MLPs to project text embeddings from the text encoder to the joint space. 
The number of layers of MLPs is a hyper-parameter that requires tuning. 
In this ablation study, the number of layers is set to $\{0,1,2,3,4,5\}$, and experiments are conducted on LVIS zero-shot evaluation with Objects365v1 and GoldG as pre-training datasets. 
The results are presented in Table \ref{ablation_study_on_text_feature_adaptor}. 
At the beginning, we observe obvious performance gains ($22.4\%\rightarrow 26.7\%$) as the number of layers increases from 0 to 3. 
Then, the performance gradually decreases when the number of layers continuously increases. 
Based on these observations, we conclude that the text feature adaptation with MLPs is significant for achieving better alignment in the joint space. 
Consequently, the number of layers of MLPs is set to 3 for all our models.

\begin{table}[h]
\caption{Ablation Studies on Text Feature Adaption. The number of layers of MLPs ranges from 0 to 5. The experiment is performed on LVIS minival for zero-shot detection.}
\label{ablation_study_on_text_feature_adaptor}
\begin{center}
\begin{tabular}{c|cccc}
\textbf{Layers of MLPs} & \textbf{AP} & $\mathrm{\textbf{AP}}_{r}$ & $\mathrm{\textbf{AP}}_{c}$ & $\mathrm{\textbf{AP}}_{f}$ \\
\hline
 0 & 22.4 & 11.7 & 20.2 & 26.3 \\
 1 & 23.3 & 16.4 & 20.8 & 26.8 \\
 2 & 25.6 & 18.1 & 23.6 & 28.7 \\
 \textit{\textbf{3}} & \textit{\textbf{26.7}} & \textit{\textbf{19.9}} & \textit{\textbf{25.1}} & \textit{\textbf{29.3}} \\
 4 & 26.5 & 18.5 & 25.2 & 29.1 \\
 5 & 26.3 & 21.6 & 25.7 & 27.7 \\
\hline
\end{tabular}
\end{center}
\vspace{-8pt}
\end{table}

\subsection{Efficiency}
\label{efficiency}
\textbf{On NVIDIA RTX 4090.} 
We utilize the tool of \textit{\texttt{trtexec}} in TensorRT 8.6.1.6 \footnote{https://developer.nvidia.com/tensorrt} to assess the latency in FP16 mode.
The models are evaluated on NVIDIA RTX 4090, the size of the input image is set as $640 \times 640$ with 3 channels.
Here, we only include the YOLO-World series for comparison due to their state-of-the-art inference speed. 
Additionally, all models are re-parameterized with 80 categories from COCO.
As shown in Table \ref{efficiency_table_4090}, the parameters of our models are marginally fewer than those of the YOLO-World models. 
The inference speed of our models is significantly faster than that of the YOLO-World models. 
Notably, the FPS of the DOSOD-S model is increased by 57.1\% and 29.6\% compared to YOLO-World-v1-S and YOLO-World-v2-S models, respectively. 
YOLO-World involves complicated top-down and bottom up fusion for multi-modality and multi-level features, inevitably increasing the computational complexity. 
Whereas, our decoupled framework relies on text feature adaptation to align features from texts and images instead of mutual fusion, resulting in a concise structure and faster inference speed.

\begin{table}[h]
\caption{Comparison of parameter quantity and inference speed on NVIDIA RTX 4090.}
\label{efficiency_table_4090}
\begin{center}
\begin{tabular}{c|cc}
\toprule
\textbf{Methods} & \textbf{Params} & \textbf{FPS} \\
\hline
YOLO-World-v1-S & 13.32 M & 1007 \\
YOLO-World-v1-M & 28.93 M & 702 \\
YOLO-World-v1-L & 47.38 M & 494 \\
YOLO-World-v2-S & 12.66 M & 1221 \\
YOLO-World-v2-M & 28.20 M & 771 \\
YOLO-World-v2-L & 46.62 M & 553 \\
\hline
DOSOD-S & 11.48 M & \makecell[c]{1582\\ (\textcolor{green}{$57.1\% \uparrow$} vs v1, \textcolor{green}{$29.6\% \uparrow$} vs v2)} \\
DOSOD-M & 26.31 M & \makecell[c]{922\\ (\textcolor{green}{$31.3\% \uparrow$} vs v1, \textcolor{green}{$19.6\% \uparrow$} vs v2)} \\
DOSOD-L & 44.19 M & \makecell[c]{632\\ (\textcolor{green}{$27.9\% \uparrow$} vs v1, \textcolor{green}{$14.3\% \uparrow$} vs v2)} \\
\bottomrule
\end{tabular}
\end{center}
\vspace{-8pt}
\end{table}

\begin{table}[h]
\caption{Inference speed comparison between YOLO-World-v2 and DOSOD on D-Robotics RDK X5.}
\label{efficiency_table_X5}
\begin{center}
\begin{tabular}{c|ccc}
\toprule
\textbf{Methods} & \textbf{FPS (1 thread)} & \textbf{FPS (8 threads)}\\
\hline
YOLO-World-v2-S (INT16/INT8) & 5.962/11.044 & 6.386/12.590 \\
YOLO-World-v2-M (INT16/INT8) & 4.136/7.290 & 4.340/7.930 \\
YOLO-World-v2-L (INT16/INT8) & 4.136/7.290 & 4.340/7.930 \\
DOSOD-S (INT16/INT8) & 12.527/31.020  & 14.657/47.328 \\
DOSOD-M (INT16/INT8) & 8.531/20.238  & 9.471/26.360 \\
DOSOD-L (INT16/INT8) & 5.663/12.799  & 6.069/14.939 \\
\bottomrule
\end{tabular}
\end{center}
\vspace{-8pt}
\end{table}

\textbf{On D-Robotics RDK X5.}
We evaluate the real-time performance of the YOLO-World-v2 models and our DOSOD models on the development kit of D-Robotics RDK X5 \footnote{http://d-robotics.cc/risingSuntFive}. The RDK X5 features 8x A55 CPUs with a frequency of 1.5 GHz and a total AI power of 10 TOPS for edge computing.
The models are re-parameterized with 1203 categories defined in LVIS \cite{c34}. We run the models on the RDK X5 using either 1 thread or 8 threads with INT8 or INT16 quantization modes.
The results are shown in Table \ref{efficiency_table_X5}.
We can observe from the table that DOSOD is nearly $2 \times$ faster than YOLO-World-v2(e.g., $5.962$ $vs$ $12.527$, $4.136$ $vs$ $8.531$). Specially, in INT8 mode, DOSOD-S achieves real-time inference speed 47.328 FPS within limited computing resource.

\section{CONCLUSIONS}
\label{conclusions}
A light-weight framework, Decoupled OSOD (DOSOD), is proposed for open-set object detection. DOSOD integrates a vision-language model (VLM) with a detector. An MLP based feature adaptor is designed to transform text embeddings into a joint space, where the detector learns region representations to achieve cross-modality feature alignment without interactions. DOSOD operates like a traditional closed-set detector during testing, effectively bridging the gap between closed-set and open-set detection. Experimental results show that DOSOD significantly enhances real-time performance while maintaining comparable accuracy. Tests on the RDK X5 development kit confirm that DOSOD could well supports real-time OSOD tasks in robotic systems with limited computational resources.

\addtolength{\textheight}{-0cm}   








\end{document}